# SCFANet: Style Distribution Constraint Feature Alignment Network For Pathological Staining Translation

Zetong Chen[1#], Yuzhuo Chen[1#], Hai Zhong[2*], Xu Qiao[1*]

[1] School of Control Science and Engineering, Shandong University, Jinan 250061, Shandong Province, China

[2] Department of Radiology, The Second Hospital of Shandong University, Jinan 250033, Shandong Province, China

#These authors contributed to the work equally and should be regarded as co-first authors.

**\*Corresponding authors**: Xu Qiao, Hai Zhong

Xu Qiao

School of Control Science and Engineering, Shandong University

Address: Qianfoshan Campus, Shandong University, 17923 Jingshi Road, Jinan 250061, Shandong Province, China

Telephone number: +86-18663759381

E-mail: qiaoxu@sdu.edu.cn

Hai Zhong

Department of Radiology, The Second Hospital of Shandong University

Address: Department of Radiology, The Second Hospital of Shandong University, No. 247 Beiyuan Road, Tianqiao District, Jinan 250033, Shandong Province, China

E-mail: 18753107255@163.com

# Abstract

Immunohistochemical (IHC) staining serves as a valuable technique for detecting specific antigens or proteins through antibody-mediated visualization. However, the IHC staining process is both time-consuming and costly. To address these limitations, the application of deep learning models for direct translation of cost-effective Hematoxylin and Eosin (H&E) stained images into IHC stained images has emerged as an efficient solution. Nevertheless, the conversion from H&E to IHC images presents significant challenges, primarily due to alignment discrepancies between image pairs and the inherent diversity in IHC staining style patterns. To overcome these challenges, we propose the Style Distribution Constraint Feature Alignment Network (SCFANet), which incorporates two innovative modules: the Style Distribution Constrainer (SDC) and Feature Alignment Learning (FAL). The SDC ensures consistency between the generated and target images' style distributions while integrating cycle consistency loss to maintain structural consistency. To mitigate the complexity of direct image-to-image translation, the FAL module decomposes the end-to-end translation task into two subtasks: image reconstruction and feature alignment. Furthermore, we ensure pathological consistency between generated and target images by maintaining pathological pattern consistency and Optical Density (OD) uniformity. Extensive experiments conducted on the Breast Cancer Immunohistochemical (BCI) dataset demonstrate that our SCFANet model outperforms existing methods, achieving precise transformation of H&E-stained images into their IHC-stained counterparts. The proposed approach not only addresses the technical challenges in H&E to IHC image translation but also provides a robust framework for accurate and efficient stain conversion in pathological analysis.

# 1. Introduction

In clinical practice, histopathological examination remains the gold standard for cancer diagnosis[1]. For patients with confirmed cancer diagnoses, timely targeted therapy significantly improves survival outcomes. Currently, immunohistochemical (IHC) staining enables the detection of specific antigens or proteins through antibody-mediated visualization[2]. By evaluating antigen expression levels, pathologists guide clinicians in administering targeted treatments. Moreover, assessing antigen expression is crucial for monitoring therapeutic efficacy and prognosis. Consequently, the rapid intraoperative and cost-effective postoperative acquisition of IHC images is of paramount importance. However, IHC staining requires specialized technicians and laboratory equipment, making it both time-consuming and expensive[3]. These limitations hinder the widespread application of IHC staining in histopathology.

Given the cost-effectiveness, shorter processing time, and operational simplicity of Hematoxylin and Eosin (H&E) staining, recent advancements have focused on leveraging deep learning-based image translation techniques to convert H&E-stained pathological sections into IHC-stained counterparts[4]. Nevertheless, inherent morphological inconsistencies and alignment errors during slide preparation introduce registration discrepancies between paired H&E and IHC images. Therefore, the translation from H&E to IHC images necessitates the development of "weakly" supervised image transformation models. Current style transfer models, which facilitate the translation of unregistered images, can be categorized into two primary classes. The first class of models utilizes content images and multiple style images during training, requiring both content and target style images during inference. Notable examples include AdaIN[5], StyleGAN[6], Styleformer[7] and DRIT[8]. The second class of models employs content images and a single style of target images during training, enabling target style transformation with only content images during inference. Prominent models in this category include CycleGAN[9], CUT[10] and UNIT[11].

Given that the core objective of this study is to perform image translation from H&E-stained images to IHC-stained images, the inference process must rely solely on H&E images without requiring target-style IHC images as input. Consequently, the first category of style transfer models, which necessitates target style images during inference, does not align with the fundamental requirements of this task. While the second category of unpaired models satisfies the basic requirements, their inherent design limitations make them less effective in accurately translating images to diverse target styles. Thus, existing style transfer models cannot be directly applied to our task.

To address these limitations, conditional guidance models incorporate the target style category as a conditional input to guide the translation process. However, in our task, the style category of the IHC image (HER2 expression level) is unknown during inference, rendering such models unsuitable for this study. These challenges prompted us to reconsider the problem: How can we design a method that, under the inherent constraints of weakly unpaired translation, neither requires target style images nor their categories as conditions during inference, yet still achieves accurate image translation with diverse target styles?

In scenarios where source and target images are perfectly aligned, the strongly supervised image translation model Pix2Pix[12] has demonstrated promising results. This model employs L1 loss and GAN loss[13], where the GAN loss ensures that the generator's output distribution aligns with the target distribution, and the pixel-level L1 loss constrains the generated image's pixel positions and overall style to match those of the target image. Although Pix2Pix is not directly applicable to our weakly unpaired task, it provides valuable inspiration: If we can devise alternative methods to constrain the style and structure of the generated images, enabling their application in weakly unpaired scenarios, we can achieve image translation without relying on target style images or their categories during inference.

Therefore, we propose the Style Distribution Constrainer (SDC), which not only ensures that the generator's output distribution aligns with the overall target image distribution through GAN loss but also constrains the generated image to match the style of weakly unpaired target images. Additionally, we employ Cycle Loss to maintain structural consistency between the generated image and the source image. We compare this approach with conventional methods based on Perceptual Losses[14], which are commonly used for style and structural constraints.

Furthermore, existing image translation models typically allow the model to autonomously learn feature transformations from the source to the target image. Such architectures lack guidance, making it challenging for the model to learn truly effective feature representations. To address this limitation, we introduce Feature Alignment Learning (FAL). The core idea is that if the hierarchical features of an image can be extracted and effectively used for reconstruction, these features are considered to well represent the image[15]. Based on this principle, in image translation tasks, it is sufficient to obtain target image features that can be used for reconstruction and to maximize the alignment of hierarchical features between the source and target images. This approach decomposes a complex translation task into two simpler subtasks—image reconstruction and feature alignment—significantly reducing the learning difficulty.

The main contributions of this study are as follows:

1)Our proposed model, SCFANet, addresses the weakly unpaired problem in pathological image translation caused by morphological inconsistencies and alignment errors during operations. It neither requires target style images nor their categories during inference, ensuring its applicability in real-world scenarios.

2)To handle the diverse styles of target IHC images, our proposed SDC achieves better style constraints compared to conventional style losses.

3)We introduce FAL, which decomposes a complex translation task into two simpler subtasks—image reconstruction and feature alignment—significantly reducing the model's learning difficulty.

4)Extensive experiments on the BCI dataset demonstrate that our model, SCFANet, outperforms existing methods and achieves precise translation from H&E-stained images to IHC-stained images.

## 2. Related Work

*2.1. Image-to-Image translation networks*

The objective of this study is to achieve the translation of H&E images to IHC images, specifically implemented based on Image-to-Image Translation Networks. For paired source-target image datasets, introducing a strongly supervised learning mechanism has proven to be the most effective approach. The Pix2Pix[12] model, by combining adversarial loss and pixel-level loss (L1), achieves high-quality image translation on paired source-target datasets. Relevant research in medical image cross-modal translation includes studies such as[16][17][18]. These methods significantly improve the accuracy and consistency of generated images by training on registered source-target image pairs. However, due to morphological inconsistencies and alignment errors during slide preparation, H&E-IHC images exhibit certain registration discrepancies. Consequently, the strongly supervised image translation model Pix2Pix is not suitable for this study.

To address this issue, researchers have explored various unsupervised learning methods. For instance, CycleGAN[9] employs cycle-consistency loss, while CUT[10] utilizes mutual information loss. Relevant studies in medical image cross-modal translation include[19][20][21]. These methods enable effective cross-modal image translation on unpaired data, providing more flexible and practical solutions for medical image generation. Such models use content images and a single style of target images during training, requiring only the content images during inference to achieve target style transformation. However, as shown in Fig.1, IHC images with different HER2 expression levels exhibit distinct staining patterns. Specifically, higher HER2 expression results in darker brown staining in IHC images and greater contrast with benign tissue regions, leading to significant stylistic variations. Although unpaired models meet the basic requirements of the task, their design limitations make it challenging to accurately translate images to diverse target styles in tasks involving multiple target styles.

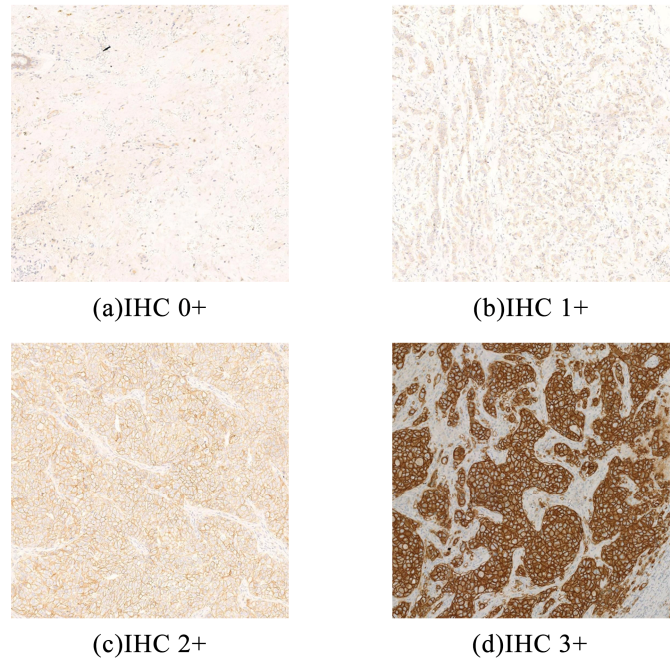

(a)IHC 0+            (b)IHC 1+

(c)IHC 2+            (d)IHC 3+

**Fig.1.** IHC images of different HER2 expression levels.

For image translation tasks involving target images with multiple styles, Classifier Guidance Diffusion Models[22], StarGAN[23] utilize conditional guidance by specifying the target style category, enabling the generator to translate content images into target images of a specific style. In our task, the target style category corresponds to the HER2 expression level, and there are two approaches to obtain this condition during model inference: (1) Obtain the IHC-stained slice paired with the H&E-stained image, evaluate the HER2 expression level from the IHC image, and use this information to guide the translation of the H&E-stained slice into an IHC-stained image[24]. (2)Predict the HER2 expression level directly from the H&E image using a classification model, and use this prediction to guide the translation of the H&E-stained slice into an IHC-stained image[4].

However, the first method requires the use of corresponding IHC-stained images during model inference, which contradicts the fundamental premise of this study. The second method introduces classification errors, and since the generated results are biased toward the given conditions, even minor classification errors can render the generated images entirely devoid of diagnostic value. Consequently, existing image translation models are ill-suited for the task of

H&E to IHC image translation in this study. To address these limitations, we propose SCFANet, which introduces the Style Distribution Constrainer (SDC) to enforce style consistency and combines it with Cycle Loss to ensure structural consistency. This approach partially resolves the weakly unpaired problem in pathological image translation caused by morphological inconsistencies and alignment errors during operations. Unlike existing methods, SCFANet neither relies on target style images nor uses their categories as conditions during inference, ensuring its applicability in real-world scenarios.

*2.2. Pathological Staining Translation*

Pathological image translation has emerged as a rapidly advancing field in medical imaging. Numerous studies have developed frameworks and algorithms for generating virtual functional staining images from conventional staining images. Huang et al.[25] proposed a Lesion-Aware Generative Adversarial Network (LA-GAN) to enhance the visual representation of lesion features in generated images. Fernandez et al.[26] focused on generating multi-pathological and multi-modal images and labels using techniques such as latent space sampling and autoencoders. Shaban et al.[27] introduced StainGAN for pathological image translation, which is based on the principles of CycleGAN. Li et al.[28] trained a convolutional neural network to establish mappings between unstained and stained tissue images using a conditional generative adversarial network model.

Among pathological image translation researches, the translation from H&E to IHC images has become one of focal points due to the ability of IHC staining to reveal the presence of specific antigens or proteins through the staining of corresponding antibodies. Liu et al.[29] proposed Pyramid Pix2Pix, a framework specifically designed for structurally aligned data, which overcomes the limitations of pixel-level alignment. Their method achieved a PSNR of 21.160 and an SSIM of 0.477 on the BCI dataset. Li et al.[30] introduced the Adaptive Supervised Patch (ASP) Loss, which directly addresses the inconsistency between input and target images in H&E-to-IHC image translation frameworks. Their approach achieved SSIM scores of 0.5236 and

0.2159 on the BCI and MIST datasets, respectively. Zhang et al.[31] proposed Multiple Virtual Functional Stain (MVFStain), a method capable of generating multiple functional stain images simultaneously from H&E-stained histopathological images. Notably, their model achieved a PSNR of 26.1919 for HER2 functional staining in breast tissue.

The premise of converting H&E to IHC images lies in the existence of a modelable implicit relationship between the two modalities. Tewary et al.[32] employed transfer learning for HER2 scoring, achieving an accuracy of 93% using VGG19. Akbarnejad et al.[33] constructed a large-scale dataset to predict Ki67, ER, PR, and HER2 statuses from H&E images, achieving an AUC of approximately 90% based on a ViT-based pipeline. Han et al.[34] utilized a reparameterization scheme to decouple training and deployment models, achieving a HER2 scoring accuracy of 94% based on H&E images. These findings suggest that while H&E images cannot directly enable pathologists to assess HER2 expression levels based on staining patterns, as is possible with IHC images, they do contain implicit high-dimensional features that allow models to evaluate protein or antigen expression. This indirectly validates the relationship between H&E and IHC. Therefore, the primary requirement for a model translating H&E images to IHC images is to transform the implicit high-dimensional features of H&E into IHC images with specific staining patterns, ensuring consistency in classification results and image style while maintaining structural integrity. To address this, the proposed SCFANet comprehensively constrains the generated images from three perspectives: style, structure, and pathological pattern consistency. Recognizing the inherent difficulty of this task and the lack of guidance in existing image translation model architectures, we further introduced FAL to achieve efficient and precise translation from H&E to IHC images.

## 3. Proposed Method

In this section, we provide a detailed description of the network architecture of the Style Distribution Constraint Feature Alignment Network (SCFANet), as illustrated in Fig.2. SCFANet incorporates two novel components: the Style Distribution Constrainer (SDC) and the Feature

Alignment Learning (FAL). The SDC ensures style consistency between the generated images and the target images, while being integrated with cycle consistency loss to enforce constraints on both the structure and style of the generated images. The FAL decomposes the complex end-to-end translation task into two simpler subtasks: image reconstruction and feature alignment. Furthermore, SCFANet guarantees pathological consistency between the generated and target images by maintaining consistency in pathological patterns and Optical Density (OD).

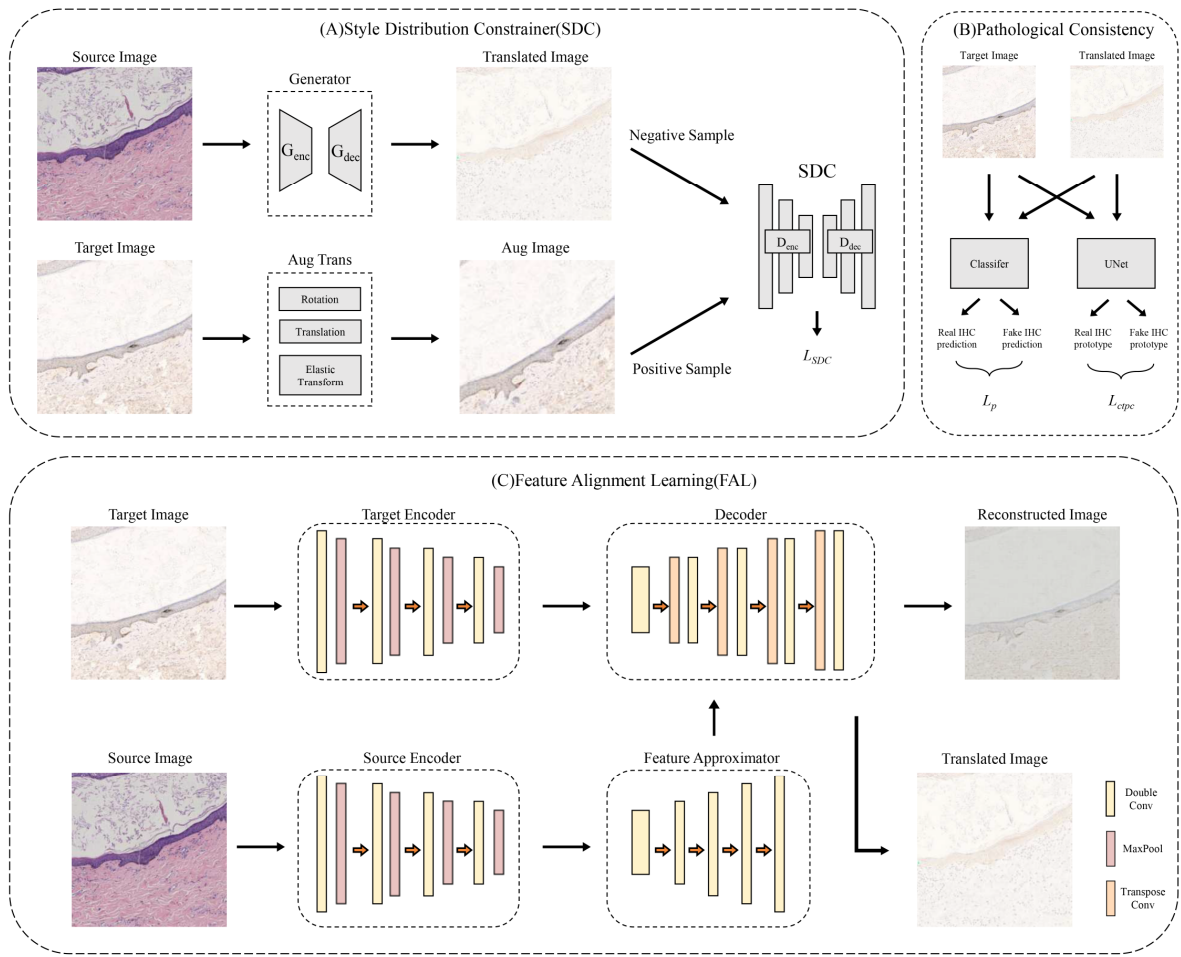

**Fig.2.** The overall framework diagram of our proposed SCFANet, which includes two novel components namely (A) Style Distribution Constrainer (SDC) and (C) Feature Alignment Learning (FAL), while guaranteeing the constraints of (B) Pathological Consistency

## 3.1. Style Distribution Constrainer

This section introduces the prerequisites and implementation details of the proposed SDC.

The Generative Adversarial Network Loss (GAN Loss) essentially aims to make the data distribution generated by the generator $P_G$ approximate the real data distribution $P_{data}$. The optimization objective can be formulated as the following minimax problem:

$$\min_G \max_D V(A, G) = \mathbb{E}_{x \sim P_{data}}[\log A(x)] + \mathbb{E}_{z \sim P_z}\left[\log\left(1 - A(G(z))\right)\right] \quad (1)$$

where $A$ is the discriminator, $\Gamma$ is the generator, $x$ represents real data, $P_{\text{data}}$ is the probability distribution of real data, $z$ is the input data to the generator, and $P_z$ is the probability distribution of the input data. As derived in [13], the original GAN Loss minimizes the Jensen-Shannon Divergence (JSD) between the generated distribution and the real distribution:

$$JS(P_{data} \| P_G) = \frac{1}{2}KL\left(P_{data} \| \frac{P_{data}+P_G}{2}\right) + \frac{1}{2}KL\left(P_G \| \frac{P_{data}+P_G}{2}\right) \quad (2)$$

The entire cluster of IHC images can be considered to reside within a broad overall distribution. The GAN Loss constrains the generated images to this distribution. However, due to variations in cell shape, size, and position, as well as differences in protein expression levels across cells, IHC images exhibit diverse staining patterns, resulting in multiple distinct styles. Thus, merely constraining the generated images to the overall distribution is insufficient for achieving precise translation from H&E images to target-style IHC images. Within the overall distribution of IHC images, clusters of images with slight deformations and displacements can be considered to reside in nearby distributions, while images with inherently different styles lie farther apart, as illustrated in Fig.3. To address this, we propose the SDC to impose style distribution constraints on the generated IHC images. This ensures that the generated images not only conform to the broad overall distribution of IHC image clusters but are further constrained to the specific small distribution corresponding to the target style.

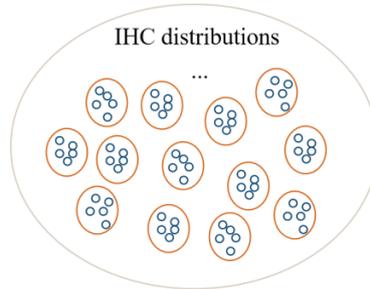

**Fig.3.** IHC coarse and fine distribution diagrams. The outermost gray circle represents the IHC coarse distribution, the orange circle inside the gray circle represents the fine distribution of the differences in IHC protein expression and cell distribution, and the blue small circle is the position of each sample in the distribution space.

The SDC is implemented by integrating the principles of adversarial learning and contrastive learning. For the SDC, IHC images that have undergone slight deformations and displacements should be considered to reside within the same distribution. Therefore, we treat the target IHC images with minor deformations and displacements as positive samples, concatenate them with the target IHC images, and assign them a true label. Conversely, the generated IHC images are treated as negative samples, concatenated with the target IHC images, and assigned a false label. When the labels in supervised learning are non-registered gold standard images, the SDC encourages the generator to generate IHC images that retain the spatial structure of H&E images and only alter the staining style. Such images exhibit minimal deformations and displacements compared to the target images but reside within the same style distribution. In contrast, pixel-level strong supervision losses like L1 encourage the generator to produce deformed images. Without providing the generator with explicit deformation guidance, it may disrupt the inherent spatial relationships.

The SDC operates as a binary classifier, and its loss function is defined as follows:

$$L_{SDC} = \{L_{BCE}[\Sigma(I_{trg}, I_{sim}), 1] + L_{BCE}[C(I_{trg}, I_{gen}), 0]\} \times 0.5 \tag{3}$$

Here, $L_{SDC}$ represents the loss function of the SDC, $L_{BCE}$ denotes the Binary Cross Entropy Loss, and $\Sigma$ is the SDC module, which takes two input images. Specifically, $I_{trg}$ is the target image, $I_{sim}$ is a simulated unpaired image obtained by applying predefined deformations to the target image, and $I_{gen}$ is the target image generated by the generator through stain transformation of the source image. Through the design of this loss function, we treat the generated IHC images and the target IHC images as belonging to different distributions. This encourages the generator to generate images that exhibit only minor deformations and displacements compared to the target images, ensuring they reside within the same style distribution.

*3.2. Structure Constraint Loss*

In the previous section, we proposed the use of the SDC to enforce style constraints, ensuring that the generated images maintain style consistency with the target images. Simultaneously, to guarantee structural consistency between the generated images and the source images, we introduce the Cycle Loss[9] as the loss function for structural constraints. The Cycle Loss ensures cycle consistency between the source images and the generated images, and it is defined as:

$$L_{cyc} = L_1\left[I_{src}, \mathbf{\Gamma}_{rev}(I_{gen})\right] \qquad (4)$$

Here, $L_1$ presents the L1 loss, which computes the Mean Absolute Error (MAE) between two images. $I_{src}$ denotes the source image, and $\mathbf{\Gamma}_{rev}$ is the reverse generator, which takes the target image as input and transforms it back into the source image.

*3.3. Pathological Consistency Loss*

To further ensure that the model learns the pathological pattern consistency between the generated images and the target images, this study introduces a pattern classifier $\mathbf{X}$. This classifier is pre-trained on real IHC images and then integrated into the training process of the SCFANet. Specifically, the pre-trained classifier is employed to classify the generated images and the target images during training, and the cosine similarity between their predicted outputs is computed as a loss function:

$$L_p = 1 - \text{Cosine Similarity}[\mathbf{X}(I_{trg}), \mathbf{X}(I_{gen})] \qquad (5)$$

$$\text{Cosine Similarity}(A_i, B_i) = \frac{\sum_{i=1}^{n}(A_i \times B_i)}{\sqrt{\sum_{i=1}^{n} A_i^2} \times \sqrt{\sum_{i=1}^{n} B_i^2}} \qquad (6)$$

Here, $L_p$ represents the pattern loss, and *Cosine Similarity*$(A_i, B_i)$ denotes the cosine similarity between the two vectors. In addition to enhancing the model's ability to learn pathological pattern consistency, the introduced pattern classifier also serves as a regularization mechanism, helping to mitigate instability issues in generative adversarial network training, such as mode collapse. Furthermore, it enables the evaluation of the model's style translation accuracy by computing the classification accuracy of the generated images, as well as assessing whether the generated IHC images can be clinically utilized for detecting HER2 expression.

In addition to the global pathological pattern consistency loss, we incorporate the Cross-image Tumor Prototype Consistency (CTPC) Loss proposed by Johnson et al[35]. This loss ensures that the tumor content in the generated images remains consistent with that in the target images. It is computed using a focal optical density mask, which is derived from the Focal Optical Density (FOD) of the images.

### 3.4. Feature Alignment Learning

In this section, we introduce the network architecture and loss function corresponding to FAL .

#### 3.4.1. Network architecture

The architecture of the FAL framework is illustrated in Fig.4. During training, the generator simultaneously processes both the target image and the source image. The generator based on FAL adopts a dual-encoder single-decoder architecture. The two encoders share the same structure but do not share parameters. To obtain reconstructible target image features, the Target Image Encoder and the decoder form an Autoencoder structure, performing target-to-target image reconstruction. The Autoencoder is designed to learn the reconstruction of the target image from itself, ensuring its ability to capture and reproduce the target image's features. The Source Image Encoder extracts hierarchical features from source images. The Feature Approximator is responsible for learning the transformation of the hierarchical features from the source image into those of the target image, thereby facilitating feature-level alignment between the two domains.

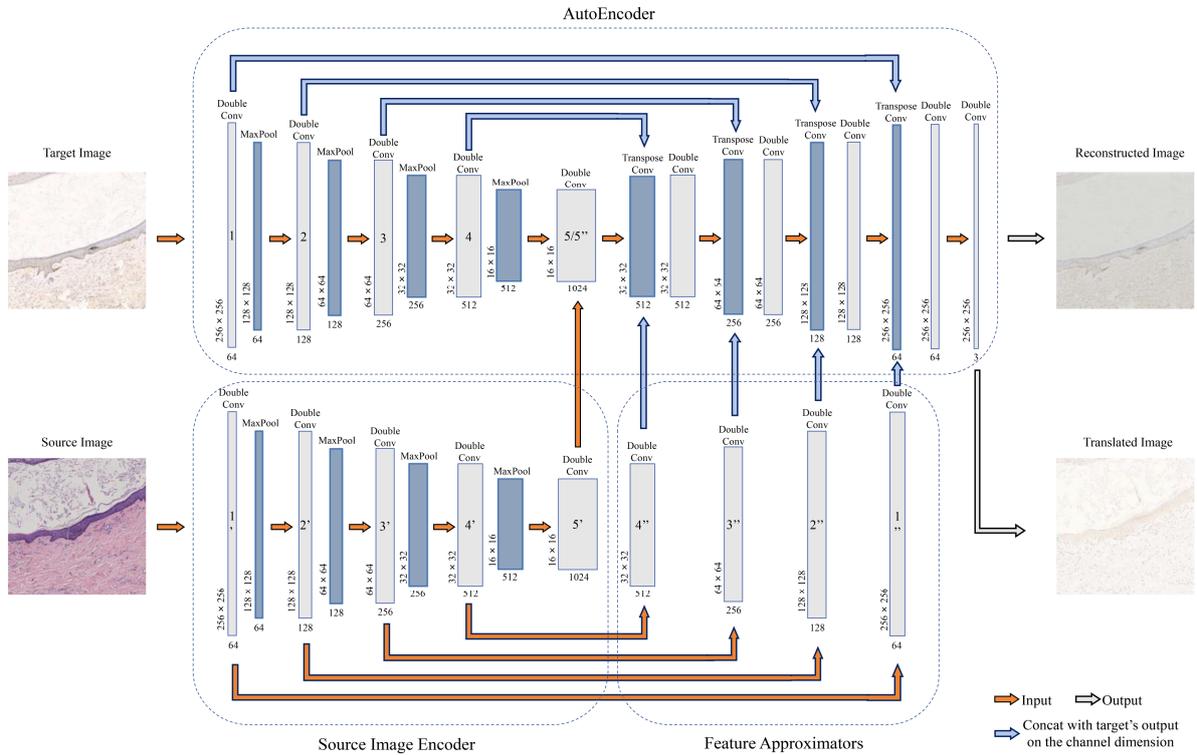

**Fig.4.** The overall view of FAL's architecture. In the figure, Double Conv1-5 is the target image encoder, Double Conv1'-5' is the source image encoder, Double Conv1"-5" is the feature approximator (Double Conv5/5" is a double-layer convolution with unshared parameters, Double Conv5 and Double Conv5"), and the unlabeled Double Conv is the feature decoder shared by the target image and the source image. Feature alignment and feature decoding parameter sharing are the most important ideas of this model.

In this framework, the encoder of the Autoencoder serves as the Target Image Encoder, which extracts five levels of features from the target image through Double Convolution (Double Con) layers 1-5. Correspondingly, the Source Image Encoder employs Double Con layers 1'-5' to extract five levels of features from the source image, aligning them with the corresponding levels of the target image. Subsequently, the Feature Approximator utilizes Double Con layers 1"-5" to transform the five levels of features from the source image into their counterparts in the target image. The transformed features from the source image are treated as approximate features of the target image and are fed into the decoder of the Autoencoder using the same logic as in the reconstruction process, thereby generating the target image from the source image.

By training an Autoencoder with high-quality reconstruction capabilities and sharing the pre-trained weights of the feature decoder, the FAL framework leverages the target image features as guidance. This enables the Source Image Encoder and the Feature Approximator to collaboratively learn, ensuring that the features of the source image at each level closely approximate those of the target image at the same level. Ultimately, the shared feature decoder generates a high-quality target image. It is important to note that during testing or inference, our method does not require the target image as input, and the Double Con layers 1-5 can be discarded. Instead, the process is simplified as follows: source image → Source Image Encoder → Feature Approximator → Feature Decoder, with the output image serving as the final result.

*3.4.2. Loss function*

The generator architecture based on FAL operates in two distinct training phases: (1)The Autoencoder phase, where the Target Image Encoder and the shared feature decoder are trained. (2)The transformation phase, where the Source Image Encoder and the Feature Approximator are trained. The loss functions vary across these phases and are categorized into $L_{rec}$ and $L_{gen}$:

$$L_{rec} = L_{GAN}\left\{\mathbf{A}\left[\mathbf{\Delta}_\theta\left(\mathbf{E}_{trg}(I_{trg})\right)\right], \mathbf{1}\right\} + \lambda_{Cha}L_{Cha}\left[I_{trg}, \mathbf{\Delta}_\theta\left(\mathbf{E}_{trg}(I_{trg})\right)\right] \quad (7)$$

$$L_{gen} = L_{GAN}\left\{\mathbf{A}\left[\mathbf{\Delta}_{\hat\theta}\left(\mathbf{E}_{src}(I_{src})\right)\right], \mathbf{1}\right\} + \lambda_{SDC}L_{SDC} + \lambda_{cyc}L_{cyc} + \lambda_p L_p + \lambda_{cptc}L_{cptc} \quad (8)$$

Here, the discriminator $\mathbf{A}$ uses frozen parameters during the generator training phase. $\mathbf{E}_{trg}$ and $\mathbf{E}_{src}$ represent the Target Image Encoder and Source Image Encoder, respectively. $\mathbf{\Delta}_\theta$ denotes the shared feature decoder, which is trained in the first phase and uses frozen parameters in the second phase. 1 is a label matrix with the same dimensions as the output ground truth, filled with ones. $L_{Cha}$ is the Charbonnier Loss[36]. The computational formula is shown in Equation (9), which is a simple variant of the L1 loss. The Charbonnier loss is introduced in this study to avoid excessive smoothing of the generated image. $\lambda_{Cha}$、$\lambda_{SDC}$、$\lambda_{cyc}$、$\lambda_p$、$\lambda_{cptc}$ are the weighting coefficients for their respective losses.

$$L_{Cha}(x,y) = \frac{\sum_{i,j,k}\sqrt{(x_{i,j,k}-y_{i,j,k})^2+\epsilon^2}}{H\times W\times C} \quad (9)$$

where $x, y$ are two images with $H \times W \times C$ dimension .

For the discriminator, the loss function is:

$$L_A = \left\{ L_{GAN}\left\{ \mathbf{A}\left[ \mathbf{\Delta}_{\hat{\theta}}\left( \mathbf{E}_{src}(I_{src}) \right) \right], \mathbf{0} \right\} + L_{GAN}\left[ \mathbf{A}\left( I_{trg} \right), \mathbf{1} \right] \right\} \times 0.5 \tag{10}$$

In this phase, the discriminator $\mathbf{A}$ and the SDC $\mathbf{\Sigma}$ are trained simultaneously with the parameters of the other networks are frozen.

## 4. Experiment settings

*4.1. Datasets*

For the experiments in this study, the Breast Cancer Immunohistochemical (BCI) public dataset was selected[4]. This dataset is a collection of breast cancer immunohistochemically stained pathological images and represents the first publicly available dataset specifically designed for generating breast cancer IHC-stained images. The dataset was constructed by sequentially slicing two layers from the same tumor tissue, staining them with H&E and IHC respectively, and then scanning the prepared pathological slides into Whole Slide Images (WSI). The BCI dataset utilized the Hamamatsu NanoZoomer S60 scanner to capture both H&E-stained WSIs and their corresponding IHC-stained WSIs. Subsequently, the downsampled H&E-IHC WSI pairs were aligned using projective transformation and elastic registration. Finally, the WSIs were segmented into square patches with a side length of 1024 pixels, and regions that did not contain tumor tissue or failed to align through the two-step registration process were filtered out. The BCI dataset comprises 9746 images (4873 pairs), with 3896 pairs allocated for training and 977 pairs for testing, encompassing a wide range of HER2 expression levels. Representative samples of H&E-IHC image pairs are illustrated in Fig.5.

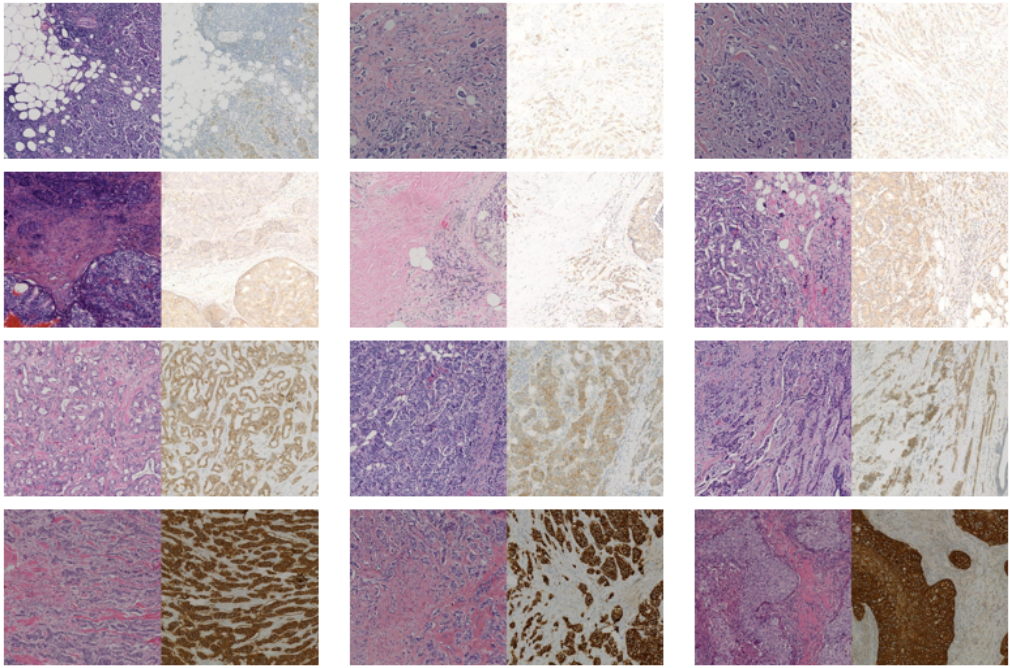

**Fig.5.** Some H&E-IHC image pairs of BCI

*4.2. Experiment details*

In this study, the PyTorch deep learning framework was employed in conjunction with the Python 3.8 interpreter. The model was trained using an A100 GPU with 80GB of memory, and the CUDA version utilized was 11.8. The Adam optimizer was selected for training all networks. During the training of the generative network, the input images were subjected to a series of random transformations, including elastic deformation with intensity ranges of [5, 100], translation with scaling ranges of [-0.2, 0.2], and rotation with angle ranges of [-15°, 15°]. These transformations were randomly combined to enhance the robustness of the model. To prevent the generation of black borders in the images, the maximum bounding box was calculated for each transformation, and the images were cropped accordingly. Additionally, to improve training stability, the pixel values of the images were normalized to the range of [-1, 1]. During testing, each image was resized to a resolution of 256×256, and the pixel values were similarly normalized to the range of [-1, 1]. To ensure intensity consistency between the source and target images and to avoid potential biases in structural and content alignment caused by illumination differences, illumination normalization was applied in this study. Specifically, the overall mean intensity of the source and target images was unified.

In this study, we set the batch size, the learning rate to 1 and 1e-4. The normalization method is Instance Normalization, and the number of epochs are 500. $\lambda_{cha},\lambda_{SDC},\lambda_{cyc},\lambda_{p},\lambda_{cptc}$ are respectively set to 100,10,10,20,2.5.

*4.3. Model evaluation parameters*

In natural image translation tasks, higher Peak Signal-to-Noise Ratio (PSNR) and Structural Similarity Index Measure (SSIM) typically indicate superior image quality. However, this correlation does not consistently hold true for pathological image translation. Specifically, blurred generated IHC images may lead to artificially inflated PSNR and SSIM values. Furthermore, due to the weakly unpaired issue between image pairs, relying solely on PSNR and SSIM measurements cannot precisely reflect the quality of generated images. Therefore, we present SSIM and PSNR metrics as reference points rather than as fundamental criteria for evaluation.

To objectively assess the pathological consistency between generated and target images, this study introduces the Optical Density (OD) difference $L1_{OD}$ between them. The calculation formulas are as follows:

$$OD_C = \left(-log_{10}\left(\frac{I_C}{I_{0,C}}\right)\right)^{\alpha} \quad (11)$$

$$L1_{OD} = L1(OD_{trg}, OD_{gen}) \quad (12)$$

Where $I_{0,C}$ and $I_C$ represent the incident and transmitted light intensities respectively, $OD_C$ denotes the optical density of channel C, $OD_{trg}$ and $OD_{gen}$ represent the optical densities of target and generated images respectively, and $L1$ is the L1 loss. We implement this formula by converting the DAB channel of IHC to grayscale and assigning grayscale values to positive signals using a focus calibration map.

For better evaluation of visual effects, we have incorporated Visual Information Fidelity

(VIF). VIF is a metric that measures the visual information fidelity between generated and target images, incorporating characteristics of the human visual system to more accurately reflect human perception of image quality. The calculation formula is as follows:

$$VIF = \frac{\sum_{i,j} I(C_{ij}^N; E_{ij}^N | s_{ij})}{\sum_{i,j} I(C_{ij}^N; R_{ij}^N | s_{ij})} \tag{13}$$

Where $C_{ij}^N$ represents the information of the reference image at the j scale and i subband, $E_{ij}^N$ denotes the information of the generated image at the j scale and i subband, $R_{ij}^N$ indicates the information of the target image at the j scale and i subband, $S_{ij}^N$ represents the local statistical characteristics of the generated image at the j scale and i subband, and I denotes the conditional mutual information.

Additionally, to evaluate the accuracy of model style translation and assess whether the generated IHC images can be clinically used for detecting HER2 expression, we have introduced Accuracy (Acc) and Area Under the Curve (AUC) as comprehensive and objective evaluation metrics. Due to the issue of class imbalance, when averaging metrics across multiple categories, we have adopted weighted averages to ensure more objective assessment.

## 5. Results and Discussion

*5.1. Ablation experiments results*

Ablation study is a scientifically rigorous experimental design methodology that systematically adds, removes, or modifies specific factors in a study to assess their impact on outcomes and validate hypotheses. To verify and elucidate the influence of different modules in SCFANet on the original baseline CycleGAN, this study conducted ablation experiments to evaluate the contributions of CTPC Loss, SDC, FAL, and Pattern Consistency (PC) Loss.

In the ablation experiments, a multi-factor experimental approach was adopted to demonstrate the effectiveness of comprehensive improvements. On the BCI dataset, each subsequent experimental group incorporated additional improvements based on the previous group. The experimental results, as shown in Table 1, demonstrate that the introduction of each new module consistently enhanced the performance of the previous group. Notably, the model incorporating all four comprehensive improvements—the final SCFANet—achieved optimal results across all metrics: $L_{SDC}$, $L1_{IOD}$, Acc, AUC, and VIF. Specifically, $L_{SDC}$ decreased from 2.1912 to 1.3423, and $L1_{IOD}$ decreased from 16633.60 to 10299.89. Meanwhile, Acc, AUC, and VIF improved from 0.3214, 0.3214, and 0.8313 to 0.7124, 0.7811, and 0.9007, respectively.

The introduction of the CTPC module reduced $L1_{IOD}$ from 16633.60 to 13674.38, effectively validating the consistency constraint on OD, which relates to the the tumor content. The inclusion of SDC decreased $L_{SDC}$ from 2.0335 to 1.7070, demonstrating its effectiveness in constraining the style of generated images. Furthermore, the PC module significantly improved Acc and AUC from 0.4084 and 0.4084 to 0.7124 and 0.7811, highlighting its role in constraining pathological patterns, particularly HER2 expression levels, in generated images.

Although all other performance metrics improved with the introduction of new modules, PSNR and SSIM did not follow this trend. While PSNR and SSIM increased with the addition of the first module (CTPC), subsequent modules caused these metrics to decline. This observation further underscores the limitations of PSNR and SSIM in pathological image translation tasks. In summary, the ablation experiments demonstrate that each proposed module and improvement effectively enhances model performance, robustly validating the efficacy of the proposed methodology.

**Table 1**. Ablation experiments for each module in SCAFFNet. The added modules include CTPC, SDC, FAL, PC. The results show the improvement of each module on the network performance (in the table, the highest performance metrics are bolded).

| Model | PSNR | SSIM | $L_{SDC}\downarrow$ | $L1_{IOD}\downarrow$ | Acc | AUC | VIF |
|---|---|---|---|---|---|---|---|
| CycleGAN | 22.85 | 0.4523 | 2.1912 | 16633.60 | 0.3214 | 0.3214 | 0.8313 |
| CycleGAN+CTPC | **23.79** | **0.4699** | 2.0335 | 13674.38 | 0.3398 | 0.3398 | 0.8541 |
| CycleGAN+CTPC+SDC | 23.37 | 0.4442 | 1.7070 | 11840.78 | 0.3849 | 0.3849 | 0.8767 |
| CycleGAN+CTPC+SDC+FAL | 22.86 | 0.4042 | 1.3978 | 10958.01 | 0.4084 | 0.4084 | 0.8852 |
| CycleGAN+CTPC+SDC+FAL+PC | 22.75 | 0.3938 | **1.3423** | **10299.89** | **0.7124** | **0.7811** | **0.9007** |

*5.2. Different structure and style constraint losses comparison performance*

This study introduces the SDC, which not only ensures that the generated images conform to the overall distribution of the target images through GAN loss but also further constrains the style consistency between the generated and target images. Additionally, Cycle loss is employed to enforce structural consistency between the generated images and the source images. To validate the effectiveness of the proposed SDC in style constraint and the Cycle loss in structural constraint, we compared them with commonly used Perceptual Losses for both style and structure. Specifically, for style constraint, we used Perceptual Style Losses (PSL) and SDC as style losses, respectively, in addition to the GAN loss. For structural constraint, we employed Perceptual Content Losses (PCL) and Cycle loss as structural losses, respectively. Comprehensive comparisons were made by combining different style and structural constraints. Furthermore, to verify the limitations of strongly supervised generative models on the weakly unpaired data in this study, we also included Pix2Pix as a reference, which uses L1 loss for strong constraints on both style and structure. The experimental results, as shown in Table 2, indicate that, with the same style loss, models using Cycle loss outperform those using PCL. Similarly, with the same structural loss, models using SDC surpass those using PSL. The model combining SDC and Cycle loss achieves the best performance, confirming the effectiveness of SDC in style constraint and Cycle loss in structural constraint. Moreover, the suboptimal performance of Pix2Pix, which employs L1 loss for strong constraints on both style and structure, demonstrates that strongly supervised generative models are not well-suited for weakly unpaired data.

**Table 2**. Comparison experiments using PSL, SDC as style loss, PCL, Cycle Loss as structural loss. (Pix2Pix is introduced in the first row as a reference comparison)

| Style Loss | Structure Loss | PSNR | SSIM | $L_{SDC}\downarrow$ | $L1_{IOD}\downarrow$ | Acc | AUC | VIF |
|---|---|---|---|---|---|---|---|---|
| GAN+L1 | L1 | **24.11** | **0.4714** | 2.2084 | 12184.12 | 0.3265 | 0.3265 | 0.8623 |
| GAN+PSL | PCL | 23.09 | 0.3903 | 2.3830 | 14582.31 | 0.3490 | 0.3490 | 0.8307 |
| GAN+PSL | Cycle Loss | 22.83 | 0.4469 | 2.2212 | 13278.51 | 0.3685 | 0.3685 | 0.8764 |
| GAN+SDC | PCL | 23.53 | 0.4455 | 1.8137 | 12811.56 | 0.3746 | 0.3746 | 0.8719 |
| GAN+SDC | Cycle Loss | 23.37 | 0.4442 | **1.7070** | **11840.78** | **0.3849** | **0.3849** | **0.8767** |

*5.3. Style Distribution Constrainer strategy comparison*

The implementation of the SDC can be categorized into two approaches: pre-training and adversarial learning. The adversarial learning approach has been detailed in Section 3.1. The pre-training approach involves training an SDC independently prior to its use, rather than co-training it with the generator. During pre-training, the SDC also adopts the concept of contrastive learning, where positive samples are images subjected to minor deformations, and negative samples are images with different staining styles. The specific operations of concatenation and label assignment are consistent with those in the adversarial training described in Section 3.1, and the SDC similarly functions as a binary classifier. When training the generator, the pre-trained SDC is loaded to compute the loss $L_{SDC}$, which serves as a constraint to encourage the generator to generate IHC images that retain the spatial structure of H&E images and only alter the staining style with low $L_{SDC}$. The experimental results, as shown in Table 3, indicate that the adversarial learning approach outperforms the pre-training approach. Consequently, the adversarial learning method is adopted for the implementation of the SDC in this study.

Table 3. The model performance difference between Pre-training SDC and Adversarial learning SDC

| Model | SDC learning strategy | PSNR | SSIM | $L_{SDC}\downarrow$ | $L1_{IOD}\downarrow$ | Acc | AUC | VIF |
|---|---|---|---|---|---|---|---|---|
| CycleGAN+CTPC+SDC | Pre-training | 23.00 | **0.4495** | 1.8015 | 12935.61 | 0.3818 | 0.3818 | 0.8603 |
| CycleGAN+CTPC+SDC | Adversarial learning | **23.37** | 0.4442 | **1.7070** | **11840.78** | **0.3849** | **0.3849** | **0.8767** |

*5.4. Results of comparison with recent advances*

To further demonstrate the superiority of our proposed method, we conducted a quantitative comparison with other common approaches for H&E to IHC image translation, including CycleGAN[9], Pix2Pix[12], CUT[10], and Pyramid Pix2Pix[29]. It is worth noting that Pix2Pix and Pyramid Pix2Pix are not well-suited for weakly unpaired data in this task, while CycleGAN and CUT struggle to accurately translate images to diverse target styles in tasks with multiple target styles. All comparative experiments were performed on the BCI dataset, and multiple evaluation metrics were employed for a comprehensive comparison. The results are presented in Table 4.

The results indicate that our proposed method exhibits superior effectiveness, achieving a PSNR of 22.75, an SSIM of 0.3938, a $L_{SDC}$ of 1.3423, a $L1_{IOD}$ of 10299.89, an Acc of 0.7124, an AUC of 0.7811, and a VIF of 0.9007 on the BCI test set. Except for PSNR and SSIM, which are not suitable for this task, all other metrics reached state-of-the-art, confirming the efficacy of SCFANet.

Table 4. Comparison of the proposed SCFANet with existing methods in BCI datasets.

| Model | PSNR | SSIM | $L_{SDC}\downarrow$ | $L1_{IOD}\downarrow$ | Acc | AUC | VIF |
|---|---|---|---|---|---|---|---|
| CycleGAN | 22.85 | 0.4523 | 2.1912 | 16633.60 | 0.3214 | 0.3214 | 0.8313 |
| Pix2Pix | **24.11** | **0.4714** | 2.2084 | 12184.12 | 0.3265 | 0.3265 | 0.8623 |
| CUT | 17.29 | 0.4321 | 5.7517 | 31200.35 | 0.3327 | 0.3327 | 0.7472 |
| Pyramid Pix2Pix | 17.52 | 0.4004 | 5.8247 | 26449.13 | 0.3777 | 0.3777 | 0.7739 |
| SCFANet | 22.75 | 0.3938 | **1.3423** | **10299.89** | **0.7124** | **0.7811** | **0.9007** |

To comprehensively evaluate the performance of the proposed SCFANet model, this study not only conducted quantitative analysis but also performed visual comparisons with other methods on the BCI dataset as qualitative assessment. The experiment utilized the benchmark images from the BCI competition as the evaluation standard, and the comparison models included those featured in the BCI competition[4], ensuring the fairness of the experiment. As shown in Fig.6, the IHC images generated by our proposed SCFANet exhibit superior visual quality compared to existing methods. Specifically, for the image with HER2 expression levels greater than 3, only Pix2Pix and SCFANet were able to generate brown-stained images that meet the pathological diagnostic standards. Furthermore, compared to Pix2Pix, SCFANet produces brown-stained images that are more consistent with the target images and exhibit higher fidelity, which fully validates the effectiveness of SCFANet in constraining the style of generated images and its capability in pathological pattern constraints. Additionally, it is evident from the figure that the images generated by SCFANet not only align well with the overall structure of the source images but also maintain consistency with the tumor content in the target images, confirming the effectiveness of SCFANet in structural and pathological constraints. Integrating the quantitative and qualitative experimental results, this study robustly demonstrates the superiority and effectiveness of SCFANet in the task of H&E to IHC image translation, laying a solid technical foundation for subsequent clinical applications.

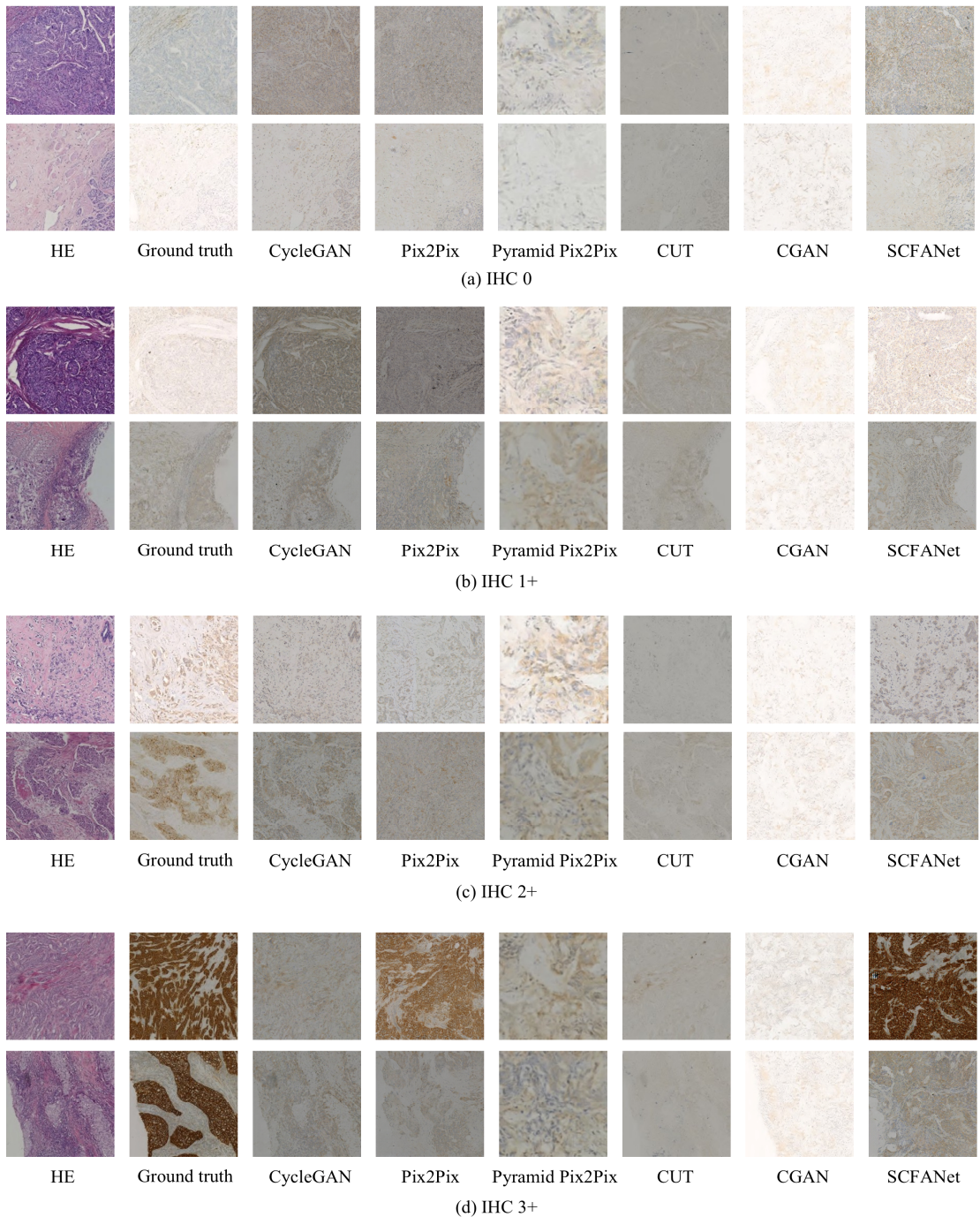

**Fig.6.** Comparative visualization of the generated images of the proposed SCFANet with common methods for H&E to IHC image translation for some data in the BCI dataset. Where (a)-(d) denote the IHC images corresponding to

different HER2 expressions

*5.5. Limitations and Extensions*

The primary objective to propose SCFANet of this study is to realize H&E to IHC image translation, yet its applications and potential extend far beyond this scope. The two core ideas of SCFANet, SDC and FAL, can be adapted to various other scenarios. SDC represents a novel approach to style constraint, ensuring that the generator's output distribution aligns with the overall distribution of the target images while further enforcing style consistency between the generated and target images. Unlike models such as CycleGAN, which rely solely on content images during inference to achieve target style translation, SDC exhibits significant potential for tasks involving multi-style image staining translation. On the other hand, FAL has even broader applicability, as it fundamentally introduces a new generator architecture. By decomposing a challenging translation task into two subtasks—image reconstruction and feature alignment—FAL significantly reduces the learning complexity. The proposed paradigm demonstrates promising potential to advance the field of image translation. However, further research is required to explore this approach and gain a more comprehensive understanding of its feasibility.

Despite the achievements of SCFANet, certain limitations remain. Firstly, the style distribution constraint of SDC relies on the assumption that the generated distribution has already converged to the overall distribution of the target images, implying that the GAN loss constraint is a prerequisite for using SDC. Additionally, as an auxiliary module within the GAN framework, SDC inevitably increases the model's complexity. Similarly, the dual-encoder single-decoder architecture of FAL faces analogous challenges. Although SDC and FAL show potential for extension to other applications, the comparative and ablation experiments in this study are confined to H&E to IHC image translation, leaving their generalizability unverified. Therefore, future research will investigate the effectiveness of SDC and FAL in other medical image translation tasks, as well as natural image translation tasks. Furthermore, we plan to optimize the implementation of SDC and FAL to reduce inference time and model parameters, thereby enhancing their scalability.

## 6. Conclusion

This paper introduces a novel framework, SCFANet, designed to achieve precise translation from H&E to IHC images. SCFANet incorporates two innovative core ideas: the SDC and FAL. SDC ensures style consistency between the generated and target images by constraining their style distribution, and it is integrated with Cycle Loss to enforce both structural and stylistic constraints on the generated images. FAL simplifies the complex end-to-end translation task by decomposing it into two subtasks: image reconstruction and feature alignment. Additionally, SCFANet enhances pathological consistency between the generated and target images by maintaining consistency in pathological patterns and OD. Extensive comparative and ablation experiments conducted on the BCI dataset demonstrate that our proposed SCFANet outperforms existing methods, achieving accurate and reliable translation from H&E to IHC images.

## Declaration of competing interest

The authors declare that they have no known competing financial interests or personal relationships that could have appeared to influence the work reported in this paper.

## CRediT authorship contribution statement

**Zetong Chen:** Writing – review & editing, Writing – original draft, Visualization, Validation, Supervision, Software, Resources, Project administration, Methodology, Investigation, Formal analysis, Data curation, Conceptualization. **Yuzhuo Chen:** Writing – original draft, Visualization, Validation, Supervision, Software, Resources, Project administration, Methodology, Investigation, Formal analysis, Data curation, Conceptualization.

**Hai Zhong:** Resources, Funding acquisition. **Xu Qiao:** Writing – review & editing, Supervision, Resources, Project administration, Funding acquisition.

## Acknowledgements

The authors would like to thank all the reviewers who participated in the review. We also acknowledge the BCI dataset for providing their meaningful datasets.

## Data Availability

Data will be made available on request.